\documentclass[runningheads]{llncs}

\usepackage{eccv}

\usepackage{eccvabbrv}

\usepackage{graphicx}
\usepackage{booktabs}
\usepackage{microtype}
\usepackage{subcaption}
\usepackage{bm} 
\usepackage[normalem]{ulem}
\usepackage{soul}
\usepackage{svg}

\usepackage[accsupp]{axessibility}  
\usepackage{makecell}
\usepackage{tabularx}
\usepackage[table]{xcolor}
\usepackage{array}
\usepackage{multirow} 
\newcolumntype{C}{>{\centering\arraybackslash}X}

\usepackage{hyperref}

\usepackage{orcidlink}

\begin{document}

\title{Test-Time Noise Guided Adaptation for Realistic Autoregressive Video Generation} 

\titlerunning{Test-Time Noise Guided Adaptation for Realistic Video Generation}

\author{Dimitrios Karageorgiou\inst{1, 2}\orcidlink{0009-0007-4259-8249} \and
Symeon Papadopoulos\inst{1}\orcidlink{0000-0002-5441-7341} \and
Ioannis Kompatsiaris\inst{1}\orcidlink{0000-0001-6447-9020} \and
Efstratios Gavves\inst{2}\orcidlink{0000-0001-8947-1332}}

\authorrunning{Karageorgiou et al.}

\institute{Information Technologies Institute, CERTH, Thessaloniki, Greece \and
University of Amsterdam, Amsterdam, The Netherlands \\
\email{\vspace{.8em}\{dkarageo,papadop,ikom\}@iti.gr, \{d.karageorgiou,e.gavves\}@uva.nl}}

\maketitle

\begin{abstract}
  Autoregressive video diffusion models have enabled the generation of arbitrarily long videos by removing conditioning on future frames, thus greatly improving computational efficiency. Yet, they suffer from error accumulation over time, as the denoised sequence gradually drifts away from the conditioning distribution seen during training. Recent advances attempt to reduce this error by anchoring each generated frame to the learned manifold of real ones. However, even when all generated individual frames lie close to the real manifold, there are trajectories which the model lacks sufficient knowledge to continue without exiting it, thus reaching a \textit{terminal point}. To prevent the model from being trapped in terminal points, we start from the hypothesis that for well-modeled future trajectories the distribution of the predicted noise should match the one of the forward noising process. To enforce such a prior at test time, we introduce \textit{Terminal points Avoidance through Noise Guided Optimization (TANGO)}, which uses the diffusion model as a critic of its own outputs, by predicting one step forward and requiring an isotropic Gaussian noise prediction. We use the deviation from this expected noise distribution to search for an alternative trajectory that does not lead to a terminal point. Our approach achieves a $3.1\%$ absolute improvement on VBench over state-of-the-art, while reducing Fr\'echet Video Distance by $28.3\%$ on average across $15$\,s videos. Our code is available on \href{https://mever-team.github.io/tango}{https://mever-team.github.io/tango}.\looseness=-1
\end{abstract}
\section{Introduction}
\label{sec:intro}

Autoregressive video diffusion models~\cite{yin2025slow, xie2025progressive, huang2025self, liu2025rolling} have recently emerged as a paradigm to generate arbitrarily long videos. To achieve this, they employ causal attention structures that condition only on past frames to generate future ones -- in practice on a limited number of them through rolling cache mechanisms~\cite{yin2025slow}. Previously, bi-directional diffusion models~\cite{xing2024survey, wan2025wan, ali2025world} had to process all frames at once, while typically operating on fixed lengths considered during training. This was limiting the maximum attainable length, as the computational cost of commonly employed architectures, such as the UNets~\cite{ronneberger2015u} and diffusion transformers~\cite{peebles2023scalable}, scales quadratically with respect to length. \looseness=-1

\begin{figure}
   \centering
   \includeinkscape[width=0.9999\columnwidth]{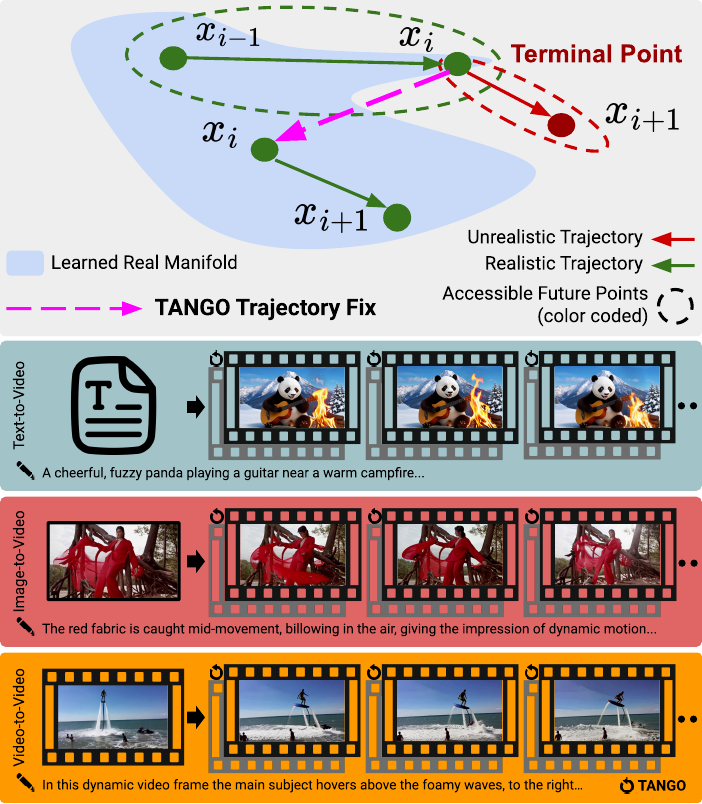}
        
    \caption{While terminal points reside in the learned real manifold, no continuation from them exists inside it. Thus, they act as traps that force the generated trajectory to continue in unrealistic regions. By employing TANGO, we avoid terminal points at test time through noise guided optimization, thus keeping the generative process within the real manifold. The upper half of the figure visualizes the trajectory correction performed by TANGO. As TANGO optimizes the visual trajectory of an autoregressive diffusion transformer by predicting a step forward, without relying on the initial condition, it can be uniformly applied for text, image and video-to-video generation. The bottom half of the figure displays three corresponding examples.}
    \label{fig:overview}
\end{figure}

While such advances enabled the generation of long videos, autoregressive approaches suffer from a training-test-time mismatch in the distribution of the past visual sequences they attend to, as typically, they are trained on a set of past trajectories originating from real videos. Instead, at inference, they have to attend to their own generated frame sequences. As the generative process progresses, small shifts in the distribution of the generated sequences from the one seen during training accumulate, causing the autoregressive process to eventually collapse~\cite{huang2025self}. We distinguish two reasons that may cause the generated sequence of an autoregressive diffusion model to shift away from its training distribution. The first -- which is the one discussed in recent literature -- is related to the imperfect noise removal during the generation of individual future frames. The denoising procedure, due to its inherent aleatoric uncertainty, as well as due to imperfections during the modeling of the training distribution, fails to match the noise levels of the training data. Therefore, denoised frames end up outside of the learned manifold of real ones. Recently, several techniques for tackling this issue have emerged. They include the temporal unrolling of the generation process during training~\cite{huang2025self, zhang2025blockvid}, the repetition of the denoising timesteps during test time~\cite{jang2026self}, or the incorporation of attention sinks~\cite{liu2025rolling, yang2025longlive, cui2026lol} to anchor the generation manifold close to the initial frames. Yet, such approaches assume that the generated sequence will match the training distribution given all its individual frames reside in the learned manifold of real ones. 

Instead, in this work we posit that even when all individual frames of a generated trajectory originate from the learned manifold of real samples, their joint distribution can differ from the one seen during training, i.e., the autoregressive process eventually generates a trajectory, for which, assuming a finite training set, no continuation exists under the learned manifold -- thus, reaching a \textit{terminal point}. Forcing the autoregressive process to continue causes it to exit the learned manifold of real samples and the produced video to become highly unrealistic. In this work, we formulate an approach to avoid terminal points and consequently, to prevent the model from generating sequences it does not know how to continue.

Understanding whether a point in the generated trajectory is terminal ultimately requires capturing the out-of-distribution conditioning sequences of the autoregressive diffusion model. To this end, our key proposition is that for well-modeled conditioning sequences, the distribution of the predicted noise should match the one considered during the forward noising process. We exploit this observation to use the denoising model itself as a critic of its past outputs, by observing whether the attributes of the predicted noise match the ones of an isotropic Gaussian distribution. To guide the diffusion model away from terminal points at test time, we introduce the Terminal points Avoidance through Noise Guided Optimization (TANGO), a method that searches for alternative trajectories that do not lead to terminal points. This is achieved by predicting one step forward and performing a constrained optimization in the neighborhood of the terminal point. Ultimately, the proposed solution continuously corrects the generated trajectory to avoid being trapped in points from which the generative model lacks the knowledge to continue. A summary of our work is presented in \cref{fig:overview}. \looseness=-1

Our paper features the following key contributions:

\begin{itemize}
    \item We highlight that terminal points prevent the use of many real trajectories as valid conditioning sequences for an autoregressive video generation model. 
    \item We posit that it is possible to capture terminal points in autoregressive diffusion models, by observing the attributes of the predicted noise.
    \item We introduce TANGO, a test-time noise guided adaptation approach, to avoid generating trajectories that lead to terminal points, by enabling a diffusion model to criticize its own outputs.
    \item Through noise guided optimization we achieve a $3.1\%$ absolute improvement over state-of-the-art on VBench~\cite{huang2024vbench}, while reducing Fr\'echet Video Distance by $28.3\%$ on average across $15$\,s videos.
\end{itemize}

\section{Related Work}
\label{sec:related}

\textbf{Video Diffusion Models.} 
Video generative modeling has been revolutionized by the advent of diffusion models~\cite{croitoru2023diffusion, ho2020denoising}, which have largely supplanted earlier adversarial~\cite{tulyakov2018mocogan, clark2019adversarial} and token-based autoregressive~\cite{yan2021videogpt, villegas2022phenaki} approaches. Initially built upon 3D U-Net architectures~\cite{ho2022video, singer2022make} and subsequently transitioning towards Diffusion Transformers (DiTs)~\cite{peebles2023scalable}, these models excel at generating high-fidelity, temporally consistent frames. Foundational works primarily focused on generating fixed-length clips by extending image diffusion models with temporal attention or 3D convolutions~\cite{ho2022video, blattmann2023align}. Scaling these architectures has led to powerful foundation models capable of simulating highly realistic physical world dynamics~\cite{agarwal2025cosmos, videoworldsimulators2024}. Concurrently, models like Lumiere~\cite{bar2024lumiere} have demonstrated the efficacy of Space-Time U-Nets (STUNets) to synthesize the full temporal duration in a single pass, contrasting with earlier cascaded temporal super-resolution methods. Recent state-of-the-art bi-directional diffusion models~\cite{wan2025wan, ali2025world} process the entire video sequence simultaneously, leveraging global temporal context to ensure high consistency across all frames. However, this global processing paradigm bounds the maximum sequence length that can be generated due to the quadratic computational complexity of attention mechanisms and the fixed context windows considered during training.\looseness=-1

\textbf{Autoregressive Video Generation.} 
Autoregressive video generation has emerged as a prominent paradigm, framing video synthesis as a sequence continuation task~\cite{kim2024fifo, deng2024autoregressive, chen2025skyreels}. This approach has been heavily driven by the development of highly compressive visual tokenizers~\cite{yu2023magvit} that allow foundation models to process multimodal inputs as discrete tokens~\cite{kondratyuk2023videopoet}. To maintain temporal coherence across longer sequences, recent methods like StreamingT2V~\cite{henschel2025streamingt2v} utilize specialized short- and long-term memory blocks.  Recently, Yin et al.~\cite{yin2025slow} introduced an approach for distilling a bi-directional diffusion transformer into an autoregressive one with a causal attention structure and a rolling KV cache, ultimately formulating text-, image- and video-to-video synthesis under a single pipeline, depending on the number of frames used to initialize the cache. However, autoregressive generation introduces significant challenges, most notably exposure bias. Because the model is trained to predict the next frames conditioned on real video frames, relying on its own imperfect generated frames during inference creates a mismatch with respect to its training distribution~\cite{huang2025self}. As generation progresses, minor deviations of the generated frame sequence from the training distribution compound, leading to severe error accumulation and the eventual collapse of the generated trajectory. \looseness=-1

\begin{figure}[t]
    \centering
   \includeinkscape[width=0.98\columnwidth]{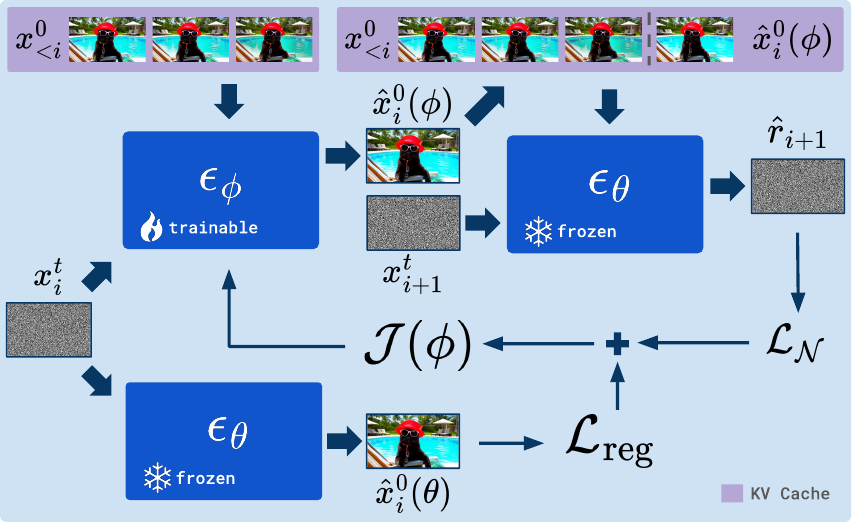}
        
    \caption{Overview of the test-time adaptation process for avoiding terminal points through noise guidance. The current output of the denoiser is criticized by moving a step forward and inspecting the properties of the predicted noise residual. Deviations from an isotropic Gaussian distribution are captured by a noise consistency objective. This objective is employed under a constrained optimization process, along with a regularization objective, to update a trainable copy of the denoiser. The output of this updated denoiser becomes the final output of each step.}
    \label{fig:architecture}
    \vspace{-12pt}
\end{figure}

\textbf{Limiting Out-of-Distribution Generated Outputs.} 
One line of work attempts to mitigate this at training time by unrolling the generation process~\cite{huang2025self, zhang2025blockvid, lu2026reward} to expose the model to its own predictions. However, this significantly increases training costs and memory overhead, while only mitigating error accumulation inside the fixed window considered during training. More efficiently, other works fine-tune noise enhancing networks to prevent the generation of low-fidelity frames~\cite{yu2025autorefiner}. Another prominent technique is the incorporation of attention sinks~\cite{liu2025rolling, yang2025longlive, cui2026lol} to anchor the generative process to the initial frames, effectively narrowing the generation manifold. Alternatively, a growing body of tuning-free inference methods attempts to maintain consistency by rescheduling noise sequences for long-range correlation~\cite{qiu2023freenoise}, applying temporal co-denoising across overlapping clips~\cite{wang2023gen}, or balancing multi-band frequency distributions during the denoising process to prevent high-frequency detail degradation~\cite{lu2024freelong}. Other approaches impose test-time objectives to align with human preference~\cite{qu2025ttom, kim2025test}, attempt to stabilize the output by repeating denoising timesteps during test time~\cite{jang2026self} so the generated frames converge to the learned real manifold or optimize the model itself to stick to the anchors~\cite{xiang2026pathwise}. Common ground in such approaches is the attempt to bring each individual generated frame into the learned manifold of real frames. When applied in autoregressive setups, they assume that if a generative model is conditioned on individual frames that resemble its training samples, it will successfully generate trajectories in the same manifold. Yet, their joint distribution is not guaranteed to match the training one, eventually leading to highly unrealistic outputs. \looseness=-1

In this paper, we argue that in order for an autoregressive model to keep generating realistic outputs over time, it is crucial for the joint distribution of its conditioning frames to not deviate from its training one, i.e. to avoid reaching terminal points. To this end, we introduce our noise guided formulation to enable an autoregressive model to capture terminal points, by inspecting the attributes of its predicted noise. Then, we employ noise guidance as a test-time optimization target to guide the diffusion model to generate trajectories without terminal points, significantly increasing the realism of the generated content, while allowing the model to explore trajectories throughout its entire learned manifold of real frames. \looseness=-1

\section{Terminal Points Avoidance Through Noise Guided Optimization}

\newcommand{\z}{\mathbf{z}}
\newcommand{\eps}{\bm{\epsilon}}
\newcommand{\thetafrozen}{\theta_{\text{frozen}}}
\newcommand{\phiadapt}{\phi_{\text{adapt}}}
\newcommand{\E}{\mathbb{E}}
\newcommand{\Loss}{\mathcal{L}}

\subsection{The Issue of Terminal Points in Generated Trajectories}

Let $\mathcal{V} = \{ x_1, x_2, \dots\}$ denote a video, where $x_i \in \mathbb{R}^{h \times w \times c}$ are its frames, and $p(\mathcal{V})$ the distribution of real videos. Under a typical autoregressive factorization~\cite{yin2025slow, huang2025self} the objective of a video generation model is to match the conditional distribution $p(x_i | x_{<i})$, i.e. to generate a plausible continuation frame given a past trajectory, as $x_i \in \text{supp}(p(x_i | x_{<i})) \Rightarrow \{x_{<i}, x_i \} \in \text{supp}(p(\mathcal{V}))$. As we consider the time horizon to be infinite, for every generated past trajectory there should be real future trajectories, thus $ \emptyset \neq \text{supp}(p(x_{i+1} | \{x_{<i}, x_i \})) \subset \text{supp}(p(\mathcal{V}))$. In practice however, an autoregressive generative model is trained on a limited set of finite length trajectories $\mathcal{D} \sim p_\mathcal{D}(\mathcal{V})$, where $\text{supp}(p_\mathcal{D}(\mathcal{V})) \subset \text{supp}(p(\mathcal{V}))$, modeling the distribution $p_{\mathcal{D}}(x_i | x_{<i})$. In this latter case, the assumption of an infinite time horizon no longer holds, so $\exists v=\{x_{<i}, x_i\}, x_i \in \text{supp}(p_{\mathcal{D}}(x_i | x_{<i}))$ such that $ \nexists u=\{v, x_{i+1}\} \in \text{supp}(p_{\mathcal{D}}(\mathcal{V})), x_{i+1} \in \text{supp}(p_{\mathcal{D}}(x_{i+1} | v))$. Consequently, even if an autoregressive generative model learns to generate perfectly valid individual future frames with respect to its training distribution, it will sooner or later generate frame sequences that jointly constitute out-of-distribution samples with respect to it. Therefore, it will be trapped in regions of the learned manifold where no valid continuation exists without exiting it. We call these regions \textit{terminal points}. A visualization of a terminal point is provided in the upper half of \cref{fig:overview}.

In order for an autoregressive model to avoid generating trajectories with terminal points it requires both a mechanism to identify them and an approach to guide it away from them. In this work we contribute to both directions. An overview of the proposed solution is provided in \cref{fig:architecture}. 

\subsection{Noise Consistency Hypothesis for Capturing Terminal Points}

From now on, we assume that the autoregressive video generation model is a diffusion one, and, for simplicity of notation and without loss of generality, we base our derivation on the score matching formulation~\cite{song2020score}. Derivations for other formulations, like DDIM~\cite{song2020denoising} or flow matching~\cite{lipman2022flow}, can be directly constructed from it -- we provide a derivation for the latter in the supplementary material.

Diffusion models are fundamentally based on the idea of moving from any random point in space along the direction that maximizes the likelihood of a sample being real, eventually reaching a point in the manifold of real samples. It has been shown~\cite{kingma2021variational, song2020score} that a model can learn to approximate this direction, starting from any point in space, by augmenting a set of real samples $x^{0}$ with isotropic Gaussian noise during the forward diffusion process:

\begin{equation}\label{eq:forward_diffusion}
    x^{t} = \alpha^t x^0 + \sigma^{t} \epsilon , \epsilon \sim \mathcal{N}(0, I)
\end{equation}

\noindent where $\alpha^t$ and $\sigma^{t}$ are scalars that control the noise level according to a noise schedule. $t$ defines the denoising timesteps indicated by the latter, and should not be confused with the temporal dimensionality of the generated video -- denoted by $i$ in the current context. Then, learning a model $\epsilon_\theta$ to remove this noise, where $\theta$ is its set of trainable parameters, ultimately approximates the score function:

\begin{equation}
    s(x^t, t) = \bigtriangledown_{x^t} \text{log}p(x^t) \approx - \frac{\epsilon_\theta(x^t, t)}{\sigma_t} = s_{\theta}(x^t, t)
\end{equation}

\noindent The denoiser is typically trained to minimize the mean squared error with respect to the added noise:

\begin{equation}
    \mathcal{L}(\theta) = \mathbb{E}_{t,x^0,\epsilon}\|\epsilon_{\theta}(x^t, t) - \epsilon\|^2_2
\end{equation}

Under a global video denoising procedure, i.e. when $x^0 = \mathcal{V} \sim p_{\mathcal{D}}(\mathcal{V})$, the only cause of $\mathcal{L}(\theta) > 0$ during testing can be the imperfect modeling of the score function due to the unavoidably limited noise samples considered during training and the limited capacity of the model. However, under an autoregressive factorization, the score function takes the following form:

\begin{equation}
    s(x_i^t, t, x^0_{<i}) = \bigtriangledown_{x^t} \text{log}p(x^t_i | x^0_{<i}) \approx - \frac{\epsilon_\theta(x^t_i, t, x^0_{<i})}{\sigma_t} = s_{\theta}(x_i^t, t, x^0_{<i})
\end{equation}

\noindent Therefore, any sequence $x^0_{<i} \notin \text{supp}(p_{\mathcal{D}}(\mathcal{V}))$ is not considered during the minimization of $\mathcal{L}(\theta)$, even if we could assume an infinite amount of noise samples and a model of adequate capacity. Thus, the output of $\epsilon_{\theta}$ is undefined in these regions and is allowed to violate the properties of isotropic Gaussian noise assumed in \cref{eq:forward_diffusion}. Overall, we exploit such violations in order to use $\epsilon_\theta$ as a critic of its own past outputs. 

We posit that at a denoising timestep $t$ it is possible to examine whether a predicted $\hat{x}^0_i$ constitutes a terminal point by examining whether the predicted look-ahead residuals $\hat{r}_{i+1} = \epsilon_{\theta}(x^t_{i+1}, t, \{x_{<i}, \hat{x}^0_i\})$ violate the properties of isotropic Gaussian noise.

\subsection{Avoiding Terminal Points by Test-Time Adaptation}

To capture deviations of $\hat{r}_{i+1} \in \mathbb{R}^{h \times w \times c}$ from isotropic Gaussian noise we compute, over its pixels, the distance of its statistical moments from the defining ones of $\mathcal{N}(0, I)$ in the spatial domain, as well as its flatness in the frequency domain.

\textbf{First and Second Moments:} Let $\mu_{\hat{r}_{i+1}}$ and $\sigma^2_{\hat{r}_{i+1}}$ be the empirical mean and biased variance of $\hat{r}_{i+1}$. We penalize deviations from the defining zero mean and unity variance of an isotropic Gaussian distribution, formulating the loss:
\begin{align}
    \Loss_{a} = \lambda_\mu(\mu_{\hat{r}_{i+1}})^2 + \lambda_\sigma (\sigma^2_{\hat{r}_{i+1}} - 1)^2
\end{align}
where $\lambda_\mu$ and $\lambda_\sigma$ are scalar hyperparameters.

\textbf{Higher-Order Moments:} To detect asymmetry and structural artifacts (heavy tails), we use the $z$-scores $z = (\hat{r}_{i+1} - \mu_{\hat{r}_{i+1}}) / \sqrt{\sigma^2_{\hat{r}_{i+1}}}$ to compute skewness and excess kurtosis. We again penalize their deviations from the values they take under an isotropic Gaussian distribution, i.e., zero: \looseness=-1

\begin{equation}
    \Loss_{b} = \lambda_\gamma\underbrace{\mu(z^3)^2}_{\text{skewness}} + \lambda_\kappa\underbrace{(\mu(z^4) - 3)^2}_{\text{excess kurtosis}}
\end{equation}

\noindent where $\mu(\cdot)$ is a pixel-wise mean operation and $\lambda_\gamma$, $\lambda_\kappa$ are scalar hyperparameters.

\textbf{Spectral Flatness:} One key property of isotropic noise is its equal energy levels in all frequencies, thus a flat power spectral density~\cite{madhu2009note}. A non-flat power spectral density would indicate spatial correlation, thus the denoiser would also be removing visual content along with the added noise. To find out whether the predicted noise residual is isotropic we evaluate the spectral flatness measure (SFM)~\cite{gray2003spectral}. Let $S(f) = |\mathcal{F}(\hat{r}_{i+1})|^2$ be its power spectral density, where $\mathcal{F}$ denotes the Discrete Fourier Transform. To ensure numerical stability and avoid log-zero errors given a zero-mean signal, we exclude the DC component ($f=0$) and add a small constant $\zeta$. Then, we define the spectral flatness loss that penalizes deviations from $1.0$ (perfect isotropic noise):
\begin{equation}
    \Loss_{s} = \lambda_s \left( 1 - \frac{\exp\left(\frac{1}{|F|}\sum_{f \in F} \ln (S(f) + \zeta)\right)}{\frac{1}{|F|}\sum_{f \in F} (S(f) + \zeta)} \right)^2,
\end{equation}
where $F$ is the set of non-zero frequencies and $\lambda_s$ a scalar hyperparameter.

\textbf{Moments in Low Frequencies:} As $\hat{r}_{i+1}$ is dominated by high frequencies, the computation of statistical moments over it is mostly overwhelmed by them. However, the energy of real visual content mostly resides in low frequencies~\cite{field1987relations}. To explicitly enforce Gaussian properties in low frequencies, we also compute skewness and excess kurtosis over a low-pass filtered variant of $\hat{r}_{i+1}$, whose $z$-scores are denoted $z^l$. We define the corresponding loss as follows:

\begin{equation}
    \Loss_{l} = \lambda_{l \mu}\mu((z^l)^3)^2 + \lambda_{l \sigma}(\mu((z^l)^4) - 3)^2
\end{equation}

\noindent where $\lambda_{l \mu}$, $\lambda_{l \sigma}$ are once again scalar hyperparameters. Then, the total noise consistency objective becomes:

\begin{equation}
    \Loss_{\mathcal{N}} = \Loss_a + \Loss_b + \Loss_s + \Loss_l
\end{equation}

To enforce this consistency at test time, we introduce a set of adaptive parameters $\phi$, initialized with the pre-trained weights $\theta$. At each autoregressive step $i$, we optimize $\phi$ to generate $\hat{x}_i^0(\phi)$ such that it satisfies $\Loss_{\mathcal{N}}$ computed using the frozen critic $\epsilon_\theta$.

To avoid trivial solutions, we formulate the adaptation as a constrained optimization problem. We update $\phi$ to minimize $\Loss_{\mathcal{N}}$ subject to a regularization constraint that anchors the search process in the neighborhood of $\hat{x}_i^0(\theta)$, which denotes the original model's prediction. The total objective becomes:

\begin{equation}
\label{eq:total_loss}
    \mathcal{J}(\phi) = \E_{\epsilon \sim \mathcal{N}(0, I)}[\Loss_{\mathcal{N}}] + \lambda_{\text{reg}} \|\hat{x}^0_{i}(\phi) - \hat{x}^0_i(\theta)\|_2^2
\end{equation}

\noindent where $\lambda_{\text{reg}}$ denotes a weighting hyperparameter. In practice, we approximate $\E_{\epsilon \sim \mathcal{N}(0, I)}[\Loss_{\mathcal{N}}]$ via a batch of $B$ noise vectors at each optimization step, controllable for the available compute power at test time. By minimizing \cref{eq:total_loss}, the optimized model is forced to avoid terminal points in the frozen manifold.

\section{Experiments}
\label{sec:eval}

\subsection{Implementation Details}

To facilitate our experiments on noise guided optimization, we follow the recent literature~\cite{liu2025rolling, yang2025longlive} and initialize our denoiser $\epsilon_\theta$ according to the distribution matching distillation scheme with self-unrolling, causal attention and KV-cache introduced by Huang et al.~\cite{huang2025self}, starting from the bi-directional video diffusion transformer~\cite{ho2022video} of Wan2.1-T2V-1.3B~\cite{wan2025wan} and using the inference optimizations of \cite{lu2026reward}. As the employed model is based on flow matching~\cite{lipman2022flow}, we solve the equation of the predicted flow with respect to noise in order to obtain the predicted noise residual $\hat{r}_{i+1}$ that we employ to guide the denoising procedure. The derivation is provided in the supplementary material. The RGB resolution of the generated video is $832 \times 480$ at $16$\,FPS and with an unlimited time horizon, due to a rolling KV cache that retains $21$ latent frames ($\approx 5$\,s of video). All the computations are taking place in the compressed latent space of a 3D variational autoencoder~\cite{kingma2013auto}, that provides $\times 8$ spatial compression and $\times 4$ temporal compression, with the exception of the initial frame that is not temporally compressed to capture richer context. Thus, $h=60$, $w=104$, while $c=16$ is used for the latent channels. During each optimization stage, we introduce a small set of trainable parameters to the self-attention layers of the causal diffusion transformer through a low-rank adaptation~\cite{hu2022lora} of rank $8$. We set $\lambda_\mu = \lambda_{\sigma} = \lambda_{\gamma} = \lambda_{\kappa} = \lambda_s = 0.1$, $\lambda_{l_{\mu}} = \lambda_{l_{\sigma}} = 0.2$ and $\lambda_{\text{reg}}=0.9$ based on a preliminary tuning on the test set of MSR-VTT~\cite{xu2016msr}, which we do not employ further in our evaluation, to avoid any potential overfitting. We optimize the model with a learning rate of $2\times 10^{-6}$ for a total of $B=10$ noise vectors, across a single epoch with batch size $2$, using the AdamW~\cite{loshchilov2017decoupled} optimizer. Experiments were conducted on a single Nvidia RTX Pro 6000 GPU with $96$\,GB of memory.

\subsection{Benchmarks \& Considered Generative Approaches}

To compute the realism of generated videos we follow recent literature~\cite{yin2025slow, huang2025self, liu2025rolling} and employ the VBench~\cite{huang2024vbench} suite. It defines a set of $946$ textual prompts, spanning $11$ evaluation dimensions and $8$ categories of visual content, which are employed to compute a set of $16$ faithfully crafted metrics. The latter are used to compute a quality and semantic score, as well as a single total score, indicative of the overall performance of a generative approach. We use all of them under a text-to-video generation pipeline. Moreover, it is not possible to account, solely through handcrafted metrics, for every possible aspect that contributes to the visual faithfulness of a video. For this reason, we also employ Fr\'echet Video Distance (FVD)~\cite{unterthiner2019fvd}, to compute the distance of the distribution of videos produced by a generative model, to the one of real videos. As a reference dataset of real videos we employ the LV-Bench~\cite{zhang2025blockvid}, which combines videos from DanceTrack~\cite{sun2022dancetrack}, GOT-10k~\cite{huang2019got}, HD-VILA-100M~\cite{xue2022advancing} and ShareGPT4V~\cite{chen2024sharegpt4v}, considering human, animal and environment-focused visual scenes, along with curated captions. To better focus on the visuals of the video, limiting the effects of a potential contextual divergence between the generated and real videos, we employ them under a video-to-video continuation pipeline. 

In our experiments, we consider a set of recently proposed autoregressive video generation approaches, with publicly available code and of similar scale. In particular, we consider CausVid~\cite{yin2025slow}, Self-Forcing~\cite{huang2025self}, RollingForcing~\cite{liu2025rolling}, LongLive~\cite{yang2025longlive} and RewardForcing~\cite{lu2026reward}, as all of them are initialized with the Wan2.1-T2V-1.3B~\cite{wan2025wan} bi-directional video diffusion transformer, thus eliminating, as much as possible, the effects of different training data and scales, enabling us to better focus on the algorithms. For a fair comparison, we use the public code of each approach to generate videos of equal length ($15$ seconds).

\begingroup
\begin{table}[t]
    \centering
    \caption{Comparison against state-of-the-art autoregressive video generation approaches on 16 visual aspects. By avoiding terminal points at test-time, TANGO enhances visual quality and semantic consistency, while better preserving content diversity. Best values are highlighted in bold, while second best are underlined.}

\scalebox{0.64}{
    \begin{tabular}{l|cccccccccc}
    \toprule
    \rowcolor{gray!20} 
    \cellcolor{white}\textbf{Approach} & 
    \makecell{\textbf{Aesthetic} \\ \textbf{Quality}} & 
    \makecell{\textbf{Appearance} \\ \textbf{Style}} & 
    \makecell{\textbf{Background} \\ \textbf{Consistency}} & 
    \textbf{Color} & 
    \makecell{\textbf{Dynamic} \\ \textbf{Degree}} & 
    \makecell{\textbf{Human} \\ \textbf{Action}} & 
    \makecell{\textbf{Imaging} \\ \textbf{Quality}} & 
    \makecell{\textbf{Motion} \\ \textbf{Smoothness}} & 
    \makecell{\textbf{Multiple} \\ \textbf{Objects}} & 
    \makecell{\textbf{Object} \\ \textbf{Class}} 
    \\
    \cmidrule(lr){1-1} \cmidrule(lr){2-2} \cmidrule(lr){3-3} \cmidrule(lr){4-4} \cmidrule(lr){5-5} \cmidrule(lr){6-6} \cmidrule(lr){7-7} \cmidrule(lr){8-8} \cmidrule(lr){9-9} \cmidrule(lr){10-10} \cmidrule(lr){11-11}
    RollingForcing~\cite{liu2025rolling} &0.658 &0.205 &0.957 &0.819 &0.403 &0.940 &\ul{0.712} &\ul{0.986} &\ul{0.932} &0.935 \\
    SelfForcing~\cite{huang2025self} &0.644 &0.206 &0.942 &0.848 &0.611 &0.950 &0.696 &0.985 &0.896 &0.959 \\
    CausVid~\cite{yin2025slow} &\ul{0.665} &\ul{0.242} &\textbf{0.966} &0.824 &\textbf{0.872} &\textbf{0.998} &0.702 &0.976 &0.747 &0.937 \\
    RewardForcing~\cite{lu2026reward} &0.650 &0.205 &0.954 &\textbf{0.896} &0.625 &\ul{0.980} &0.687 &0.984 &0.845 &0.940 \\
    LongLive~\cite{yang2025longlive} &0.660 &0.205 &0.958 &\ul{0.885} &0.361 &0.960 &0.695 &\textbf{0.988} &0.851 &\ul{0.969} \\
    \midrule
    TANGO (ours) &\textbf{0.690} &\textbf{0.257} &\ul{0.962} &0.850 &\ul{0.769} &0.960 &\textbf{0.727} &0.985 &\textbf{0.937} &\textbf{0.972} \\
    
    \end{tabular}
}

\vspace{.5em}

\scalebox{0.64}{
    \setlength{\tabcolsep}{3.2pt}
    \begin{tabular}{l|cccccc|cc|c}
    \toprule
    \rowcolor{gray!20} 
    \cellcolor{white}\textbf{Approach} & 
    \makecell{\textbf{Overall} \\ \textbf{Consistency}} & 
    \textbf{Scene} & 
    \makecell{\textbf{Spatial} \\ \textbf{Relationship}} & 
    \makecell{\textbf{Subject} \\ \textbf{Consistency}} & 
    \makecell{\textbf{Temporal} \\ \textbf{Flickering}} & 
    \makecell{\textbf{Temporal} \\ \textbf{Style}} & 
    \cellcolor{blue!15}\makecell{\textbf{Quality} \\ \textbf{Score}} & 
    \cellcolor{blue!15}\makecell{\textbf{Semantic} \\ \textbf{Score}} & 
    \cellcolor{green!15}\textbf{Total} \\
    \cmidrule(lr){1-1} \cmidrule(lr){2-2} \cmidrule(lr){3-3} \cmidrule(lr){4-4} \cmidrule(lr){5-5} \cmidrule(lr){6-6} \cmidrule(lr){7-7} \cmidrule(lr){8-8} \cmidrule(lr){9-9} \cmidrule(lr){10-10}
    RollingForcing~\cite{liu2025rolling} &0.269 &0.563 &0.747 &\textbf{0.958} &\ul{0.991} &0.241 &0.802 &\ul{0.814} &0.804 \\
    SelfForcing~\cite{huang2025self} &0.271 &0.578 &0.768 &0.928 &0.989 &0.249 &0.832 &0.808 &0.827 \\
    CausVid~\cite{yin2025slow} &\ul{0.275} &0.556 &0.667 &0.933 &0.938 &\ul{0.252} &0.837 &0.794 &0.828 \\
    RewardForcing~\cite{lu2026reward} &0.269 &0.539 &\ul{0.830} &0.932 &0.986 &0.244 &0.835 &0.804 &0.829 \\
    LongLive~\cite{yang2025longlive} &0.268 &\ul{0.591} &0.801 &\ul{0.953} &\textbf{0.992} &0.242 &\ul{0.838} &0.812 &\ul{0.832} \\
    \midrule
    TANGO (ours) &\textbf{0.283} &\textbf{0.603} &\textbf{0.859} &0.945 &0.989 &\textbf{0.275} &\textbf{0.864} &\textbf{0.860} &\textbf{0.863} \\
    \bottomrule
    \end{tabular}
}
    \label{table:sota_comparison}
\end{table}
\endgroup

\begingroup
\setlength{\tabcolsep}{9.1pt}
\renewcommand{\arraystretch}{0.9}
\begin{table}[t]
    \centering
    \caption{Ablation studies of the key losses of noise guided optimization. Average Fr\'echet Video Distance, between ground-truth and generated video segments, is reported over 3-second segments as well as over the entire videos. Lower is better. Best values in bold.}
    \scalebox{0.9}{
    \begin{tabular}{llrrrrr | r}\toprule
\multicolumn{2}{l}{\textbf{Ablation} (FVD $\downarrow$)} &\textbf{1-3s} &\textbf{4-6s} &\textbf{7-9s} &\textbf{10-12s} &\textbf{13-15s} &\textbf{Total} \\
\midrule
\midrule
\multicolumn{2}{l}{TANGO (ours)} &365.1 &\textbf{408.8} &\textbf{412.7} &\textbf{441.4} &\textbf{438.8} &\textbf{413.3} \\
\midrule
&w/o $\mathcal{L}_{a}$ &\textbf{364.6} &418.1 &467.1 &469.8 &488.2 &441.6 \\
&w/o $\mathcal{L}_{l}$ &380.3 &452.8 &518.4 &534.8 &545.2 &486.3 \\
&w/o $\mathcal{L}_{s}$ &369.3 &443.9 &507.1 &530.6 &578.4 &487.9 \\
&w/o $\mathcal{L}_{b}$ &363.7 &449.8 &561.2 &587.1 &572.3 &506.8\\
&w/o $\mathcal{L}_{\text{reg}}$ &420.1 &634.4 &630.9 & 823.9 &811.7 &664.2 \\
\midrule
&w/o $\mathcal{L}_{\mathcal{N}}$ (baseline) &367.9 &554.6 &600.7 &645.4 &715.6 &576.8 \\
\bottomrule
\end{tabular}
    }    
    \label{table:ablations}
    \vspace{-10pt}
\end{table}
\endgroup

\begingroup
\begin{table}[ht]
    \centering
    \caption{Hyperparameters ablation studies. Average Fr\'echet Video Distance, between ground-truth and generated segments, is reported for 15-second videos. Lower is better.}
    \begin{subtable}[t]{\linewidth}
\centering
\setlength{\tabcolsep}{9pt}
\renewcommand{\arraystretch}{0.85}
\scalebox{0.9}{
\begin{tabular}{cc|cc|cc}
\toprule
LoRA R. &FVD $\downarrow$ &$B$ &FVD $\downarrow$ &$\lambda_{\text{reg}}$ &FVD $\downarrow$ \\
\midrule
2 &596.2 &1 &473.7 &0.1 &656.5 \\
4 &496.3 &5 &461.8 &0.5 &449.3 \\
8 &\textbf{413.3} &10 &413.3 &0.9 &\textbf{413.3} \\
16 &425.6 &20 &\textbf{399.4} &10 &475.6 \\
\bottomrule
\end{tabular}
}
\caption{Ablations of key hyperparameters involved in test-time adaptation.}
\label{tab:hyperparams-ablation-a}
\end{subtable}

\vspace{-4pt}

\begin{subtable}[t]{\linewidth}
\centering
\setlength{\tabcolsep}{3.5pt}
\renewcommand{\arraystretch}{0.9}
\scalebox{0.9}{
\begin{tabular}{cc|cc|cc|cc|cc|cc|cc}
\toprule
$\lambda_{\mu}$ &FVD $\downarrow$ &$\lambda_{\sigma}$ &FVD $\downarrow$ &$\lambda_{l_{\mu}}$ &FVD $\downarrow$ &$\lambda_{l_{\sigma}}$ &FVD $\downarrow$ &$\lambda_{\gamma}$ &FVD $\downarrow$ &$\lambda_{\kappa}$ &FVD $\downarrow$ &$\lambda_{s}$ &FVD $\downarrow$ \\
\midrule
.01 &503.8 &.01 &482.8 &.02 &419.8 &.02 &482.8 &.01 &468.1 &.01 &479 &.01 &477.4 \\
.1 &\textbf{413.3} &.1 &\textbf{413.3} &.2 &\textbf{413.3} &.2 &\textbf{413.3} &.1 &\textbf{413.3} &.1 &\textbf{413.3} &.1 &\textbf{413.3} \\
1 &449.3 &1 &540.7 &2 &464.7 &2 &464.7 &1 &442.7 &1 &482.6 &1 &429.1 \\
\bottomrule
\end{tabular}
}
\caption{Ablations of the weights of the noise consistency objective.}
\label{tab:hyperparams-ablation-b}
\end{subtable}
    \label{table:hyperparams_ablations}
    \vspace{-16pt}
\end{table}
\endgroup

\subsection{Comparison against State-of-the-Art}

We compute all VBench metrics for all considered approaches and present them in \cref{table:sota_comparison}. Overall, we observe an absolute performance increase of $4.8\%$ in semantic score, $2.6\%$ in quality score and $3.1\%$ in the total score, as well as the best performance in most individual metrics with respect to the previous state-of-the-art. Noticeably, even in the dynamic degree metric that quantifies the diversity of generated content, our approach is only second to CausVid~\cite{yin2025slow}, which does so at the cost of semantic coherence in the generated frames. Instead, approaches that primarily rely on anchoring to the early frames to maintain visual consistency, like~\cite{yang2025longlive, liu2025rolling}, compromise their ability to generate diverse content.

\subsection{Ablation Studies}

\textbf{Loss terms:} Since our objective is to prevent the autoregressive process from exiting the learned manifold of real videos, and not to merely enforce individual visual attributes, we ablate our key choices by directly measuring the divergence of the distribution of generated videos from the one of real clips. To this end, we compute the average FVD for consecutive 3-second segments of the generated videos from their corresponding segments in the real videos of LV-Bench~\cite{zhang2025blockvid}. We report results in \cref{table:ablations}. We observe that higher-order moments are very important for capturing deviations in the predicted noise residuals from the distribution of isotropic Gaussian noise. Removing the corresponding loss impacts performance the most compared to the other losses related to noise consistency. Next, we observe that removing the objective of spectral flatness causes a significant performance drop, validating our hypothesis that spatial correlations would remove actual visual content, increasing the drift with respect to the considered distribution of real videos. Similarly, explicitly computing moments in low frequencies helps the optimization process better capture visible inconsistencies. Also, results show that regularization is crucial for preventing the optimization process from converging to trivial solutions. In fact, performing optimization without regularization degrades performance over the baseline. Overall, compared to the baseline, our test-time optimization process reduces FVD by $38.7\%$ after $15$\,s of generation, while also reducing it by $28.3\%$ on average at a video level. This indicates that our approach not only helps improve specific aspects of the generated videos, as previously presented in \cref{table:sota_comparison}, but more generally, reduces the distance between the distributions of the generated and the real videos.

\textbf{Hyperparameters:} We further ablate the key hyperparameters and report the results in \cref{table:hyperparams_ablations}. Overall, while very low LoRA ranks could even worsen performance, gains from an increase in trainable parameters eventually diminish. Instead, we observe that further increasing the number of noise vectors $B$ provides additional performance. Yet, as it comes at the cost of computational power, we set $B=10$ as a reasonable trade-off between efficiency and performance. As for the regularization coefficient and the weights of our noise consistency objective, we observe higher performance when they range in the same order of magnitude.

\begingroup
\setlength{\tabcolsep}{7.3pt}
\renewcommand{\arraystretch}{0.8}
\begin{table}[t]
    \centering
    \caption{Runtime and memory usage comparison of the proposed test-time approach with ones that enhance performance by employing a larger model.}
    \scalebox{0.9}{
    \begin{tabular}{lcccc}
\toprule
 \textbf{Model} &\textbf{Params} &\textbf{VBench} ($\uparrow$) &\textbf{Runtime} ($\downarrow$)&\textbf{Memory} ($\downarrow$) \\
\midrule
\midrule
 Baseline (SelfForcing)~\cite{huang2025self} &1.3B &0.827 &1.2 min &23.4 GiB \\
 \midrule
 HunyuanVideo~\cite{kong2024hunyuanvideo} &13B &0.832 &92.3 min &62.5 GiB \\
 SVI-1.0 (Wan2.1)~\cite{li2025stable} &14B &0.837 &30.6 min &37.9 GiB \\
 SVI-2.0 (Wan2.1)~\cite{li2025stable} &14B &0.849 &39.8 min &37.9 GiB \\
 SVI-2.0-Pro (Wan2.2)~\cite{li2025stable} &27B &0.869 &43.2 min &68.4 GiB \\
\midrule
 TANGO (ours) &1.3B &0.863 &8.6 min &37.6 GiB \\
\midrule
\bottomrule
\end{tabular}
    }    
    \label{table:runtime_analysis}
    \vspace{-14pt}
\end{table}
\endgroup

\begin{figure}
    \centering
   \includeinkscape[width=0.9999\columnwidth]{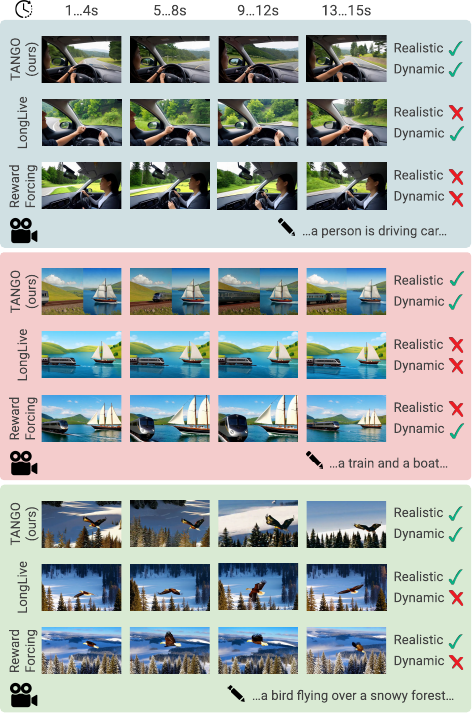}
        
    \caption{Qualitative comparison against state-of-the-art approaches.}
    \label{fig:qualitative_analysis}
\end{figure}

\subsection{Runtime \& Memory Usage Analysis}

We compared the runtime and memory usage of our test-time optimization approach against open-source video generation methods~\cite{kong2024hunyuanvideo, li2025stable} that increase visual performance primarily by scaling up the model's size. We report the respective results in \cref{table:runtime_analysis}. Overall, steering at test-time a smaller autoregressive video diffusion model~\cite{huang2025self} away from terminal points helps increase visual performance, while maintaining better trade-offs in runtime and memory usage.

\subsection{Qualitative Evaluation}

\begin{figure}[t]
    \vspace{-4pt}
    \centering
   \includeinkscape[width=0.9999\columnwidth]{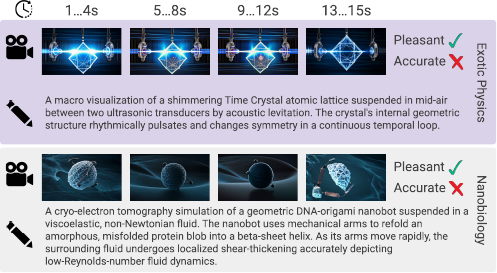}
        
    \caption{Failure cases. From left to right, frames indicative of four temporal video segments, of $15$\,s total duration, are presented. Two videos are considered, spanning the niche topics of exotic physics and nanobiology. While results are  visually pleasant, technically, they are completely inaccurate. }
    \label{fig:failure_cases}
    \vspace{-12pt}
\end{figure}

In \cref{fig:qualitative_analysis} we present a qualitative evaluation against the two best performing approaches as indicated by \cref{table:sota_comparison}. We consider indicative frames from four different segments of $15$-second videos. As shown in the first two videos, in complex scenes or scenes generated based on ambiguous textual prompts, generative models quickly get out of the learned real manifold, producing unrealistic scenes with visible artifacts. Restricting the exploration space in a narrow region around the initial points of the generated trajectory on several occasions improves realism at the cost of dynamic degree. Instead, employing TANGO to avoid being trapped in terminal points improves realism by finding a valid continuation under the limited world knowledge of the diffusion model.

\subsection{Discussion \& Failure Cases}

Any test-time adaptation approach effectively allocates inference time compute power to find a better solution under a modeled distribution, enabling the model to move away from local maxima where it may get stuck otherwise. Our approach is not an exception in this. However, as the field of large language models has shown, enabling a model to reason about its outputs~\cite{wei2022chain} and fix its mistakes is a crucial component of the inference process, with the potential of enabling models to extrapolate beyond their training regime. However, in contrast to language-based autoregressive modeling, autoregressive video diffusion models lacked a formulation to move them away from terminal points and in this work we introduce such an approach. Future work could span architectures that inherently learn to quantify their epistemic uncertainty~\cite{jazbec2025generative} without requiring gradient-descent-based optimization procedures. 

Given a fixed scale, our test-time optimization approach enhances the capabilities of a model. Yet, finding a better solution under its learned manifold requires the existence of one. For visual topics with significant underrepresentation in the training data, finding a trajectory without terminal points is challenging. Case in point, \cref{fig:failure_cases} displays the frames of two videos, with the topic of their prompts spanning exotic physics and nanobiology. While noise guidance finds trajectories that look pleasant to the eye, from a technical perspective the results are utterly wrong. Therefore, test-time adaptation is not competitive, but complementary to scaling up training data and model capacity, as the latter is the one that increases the space of valid solutions. Instead, noise guidance attempts to find a better one.\looseness=-1

\section{Conclusion}
\label{sec:conclusion}

In this work we highlighted that it is not enough for an autoregressive video generation model to merely generate future trajectories that reside in the learned real manifold, as several points in it constitute terminal points. Once reached, no continuation exists without exiting it. To avoid reaching such points, and therefore to avoid the generation of unrealistic visual content, we introduced a noise guided test-time adaptation mechanism, that uses the diffusion model itself as a critic of its own outputs. It does so by assuming that when attempting to continue a trajectory that reached a terminal point, the noise properties assumed during the forward noising process will no longer hold for the predictions of the denoiser. We exploited this observation in a constrained optimization process to move away from terminal points. Overall, by employing this test-time approach, we managed to improve state-of-the-art performance across several metrics of the realism of videos, without requiring any additional data.

\textbf{Limitations: } Any test-time adaptation approach effectively searches for an improved solution under the manifold learned during pre-training and will eventually fail if no such solution exists. Therefore, we view data and model scaling as complementary directions for increasing the space of possible solutions. Yet, our proposed noise guided optimization process contributes to finding such better solutions under the learned manifold of an autoregressive video diffusion model. \looseness=-1

\section*{Acknowledgements}
This work was supported by the Horizon Europe projects ELIAS (grant no. 101120237) and ELLIOT (grant no. 101214398). The computational resources were granted with the support of GRNET.


%
%
\bibliographystyle{splncs04}
\bibliography{main}

\clearpage

\title{Test-Time Noise Guided Adaptation for Realistic Autoregressive Video Generation\\ \mdseries{{\vspace{0.4em} \large Supplementary Material}}\vspace{-1.6em}} 

\titlerunning{Test-Time Noise Guided Adaptation for Realistic Video Generation}

\author{}

\authorrunning{Karageorgiou et al.}

\institute{}

\maketitle

\setcounter{equation}{0}
\setcounter{figure}{0}
\setcounter{table}{0}
\setcounter{section}{0}

\renewcommand{\theequation}{A\arabic{equation}}
\renewcommand{\thefigure}{A\arabic{figure}}
\renewcommand{\thetable}{A\arabic{table}}
\renewcommand{\thesection}{A\arabic{section}}

\section{Flow Matching Adaptation}\label{sec:flow_matching_adaptation}

In the main paper we formulated our terminal points avoidance through noise guided optimization process using the score matching~\cite{song2020score} derivation of generative diffusion models~\cite{croitoru2023diffusion}. This enabled us to keep notation simple and focus on our introduced test-time adaptation mechanism. Yet, several recent autoregressive video generation models~\cite{yin2025slow, liu2025rolling}, including our implementation baseline~\cite{huang2025self}, employ alternative formulations and training objectives, primarily following the optimal transport flow matching~\cite{lipman2022flow} paradigm. For the sake of reproducibility, and to facilitate the employment of our test-time adaptation mechanism to different autoregressive models, we provide below an adaptation to flow matching.

In flow matching, instead of reversing a stochastic differential equation, we construct a deterministic continuous normalizing flow that transports a base noise distribution to the data distribution. Let $x^0$ be a real sample and $\epsilon \sim \mathcal{N}(0, I)$ be isotropic Gaussian noise. We define the path $x^t$ for $t\in[0,1]$ between the data sample $x^0$ and the noise $\epsilon$. Using the optimal transport assumption~\cite{lipman2022flow}, this path is a straight-line interpolation:\looseness=-1

\begin{equation}\label{eq:forward_flow}
    x^t = (1-t)x^0 + t\epsilon, \quad \epsilon \sim \mathcal{N}(0, I)
\end{equation}

Rather than predicting the noise $\epsilon$ directly, a flow matching model $v_\theta$ (where $\theta$ represents its set of trainable parameters) is trained to regress the conditional vector field $u_t(x^t | x^0)$ that points from the data to the noise:

\begin{equation}
    u_t(x^t | x^0) = \frac{d}{dt}x^t = \epsilon - x^0
\end{equation}

The velocity model is typically trained to minimize the mean squared error against this target vector field:

\begin{equation}
    \Loss(\theta) = \E_{t, x^0, \epsilon} \| v_\theta(x^t, t) - (\epsilon - x^0) \|^2_2
\end{equation}

Under the autoregressive factorization that we consider for video generation, the target vector field incorporates the past generated history $x^0_{<i}$:

\begin{equation}
    u_t(x_i^t | x^0_{<i}, x_i^0) = \epsilon - x_i^0
\end{equation}

Thus, the trained model approximates the conditional field $v_\theta(x_i^t, t, x^0_{<i}) \approx \epsilon - x_i^0$. Similar to the score matching paradigm, any history sequence $x^0_{<i} \notin \text{supp}(p_{\mathcal{D}}(\mathcal{V}))$ is not considered during the minimization of $\Loss(\theta)$. Consequently, the predicted velocities in these regions are undefined and will violate the expected geometric constraints of a path leading to isotropic Gaussian noise instances. We exploit such violations to use $v_\theta$ as a critic of its own past outputs.

To utilize the flow matching model as a critic for terminal points, we must isolate the implied noise from the predicted velocity. Given our path in \cref{eq:forward_flow} and the velocity approximation $v_\theta \approx \epsilon - x^0$, we can algebraically express the implied noise $\hat{\epsilon}$ purely in terms of the known state $x^t$ and the model's output:

\begin{equation}
    \hat{\epsilon} = x^t + (1-t) v_\theta(x^t, t)
\end{equation}

Under an autoregressive setup, we can examine whether a predicted $\hat{x}^0_i$ constitutes a terminal point by computing its look-ahead implied noise residual $\hat{r}_{i+1}$ and examining whether it violates the properties of isotropic Gaussian noise:

\begin{equation} \label{eq:flow_matching_lookahead_residual}
    \hat{r}_{i+1} = x^t_{i+1} + (1-t) v_\theta(x^t_{i+1}, t, \{x^0_{<i}, \hat{x}^0_i\})
\end{equation}

Because $\hat{r}_{i+1}$ isolates the noise component, it can be employed as a drop-in replacement for the computation of the noise consistency objective $\Loss_{\mathcal{N}}$ (Eq. 9), presented in the main paper. Then, the test-time adaptation process presented in Sec. 3.3 is applied as is. Similar adaptations can be derived for other formulations of diffusion models~\cite{yang2023diffusion}, relying on different training objectives. Yet, it is out of the scope of this work to exhaustively cover each type of diffusion model. Instead, we focus on highlighting the importance of terminal points avoidance, which we facilitate by introducing a versatile noise guided test-time optimization mechanism. \looseness=-1

\section{Extended Implementation Details}

Following recent literature~\cite{liu2025rolling, yang2025longlive}, our autoregressive video diffusion model is initialized according to the distribution matching distillation scheme with self-unrolling, causal attention and KV-cache introduced by Huang et al.~\cite{huang2025self}, starting from the bi-directional video diffusion transformer~\cite{ho2022video} of Wan2.1-T2V-1.3B~\cite{wan2025wan}. As the latter is based on the optimal transport flow matching~\cite{lipman2022flow} formulation, we employ the adaptation presented in \cref{sec:flow_matching_adaptation} to obtain the predicted noise residual $\hat{r}_{i+1}$ that we later use in the computation of $\mathcal{L}_{\mathcal{N}}$. Along with visual conditioning, our employed diffusion transformer also conditions on latent text sequences, encoded by the umt5-xxl~\cite{chung2023unimax} pre-trained transformer.

The spatial resolution of the generated RGB video is $832 \times 480$ at $16$\,FPS and with an unlimited time horizon, due to a rolling KV cache that retains the last $21$ latent frames~\cite{liu2025rolling}, while employing the optimized inference pipeline of \cite{lu2026reward}. All the computations are taking place in the compressed latent space of a 3D variational autoencoder~\cite{kingma2013auto}, that provides $\times 8$ spatial compression and $\times 4$ temporal compression, with the exception of the initial frame that is not temporally compressed to capture richer context. Thus, $h=60$, $w=104$, while $c=16$ is used for the latent channels. Denoising is jointly performed on blocks of $3$ latent frames, i.e. $i$ in \cref{eq:flow_matching_lookahead_residual} refers to such a block. Also, four denoising timesteps are employed.

The number of initial frames in KV cache defines the type of generation. An empty cache is used for text-to-video, while single and multi-frame initializations are used for image and video-to-video accordingly. In particular, during video-to-video generation experiments we use $9$ RGB frames as a visual prompt. Additionally, in the latter case, we sample the visual prompt sequence with a framerate that matches the generator's framerate. We do this to avoid mismatches between the learned temporal representations and the ones present in the prompt video, i.e. to not slow down/speed up the prompt video when its framerate is higher/lower than $16$\,FPS. We accomplish that by discarding/repeating intermediate frames as needed.

We set the hyperparameters $\lambda_\mu = \lambda_{\sigma} = \lambda_{\gamma} = \lambda_{\kappa} = \lambda_s = 0.1$, $\lambda_{l_{\mu}} = \lambda_{l_{\sigma}} = 0.2$ and $\lambda_{\text{reg}}=0.9$ based on a preliminary tuning on the test set of MSR-VTT~\cite{xu2016msr}, under a video-to-video setup on limited-duration videos ($5$\,s), compared to our experimental evaluation, which considers videos of much greater length. The employed validation objective is the minimization of Fr\'echet Video Distance~\cite{unterthiner2019fvd} from this data. To avoid any potential overfitting, we do not further employ this data throughout our presented evaluation process.

During each optimization stage, we introduce a small set of trainable parameters to the self-attention layers of the causal diffusion transformer. We do this through low-rank adaptation~\cite{hu2022lora} of rank $8$, applied to their query, key and value matrices. We optimize the model with a learning rate of $2\times10^{-6}$ for a total of $B=10$ noise vectors on each stage, training for a single epoch with a batch size of $2$, using the AdamW~\cite{loshchilov2017decoupled} optimizer. The implementation is based on PyTorch, while FlashAttention~\cite{dao2023flashattention2} is employed for the attention operations. Gradient checkpointing was used for reducing the memory footprint during optimization. Experiments were conducted on a single Nvidia RTX Pro 6000 $96$\,GB GPU. On the latter, the generation of each RGB frame takes about $900$\,ms. 

\section{Video Length Ablation Studies}

\begingroup
\setlength{\tabcolsep}{3.9pt}
\begin{table}[t]
    \centering
    \caption{Comparison against state-of-the-art autoregressive video generation approaches on different video lengths. The combined VBench~\cite{huang2024vbench} scores for semantic consistency and visual quality, as well as the total score, are reported. Best values highlighted in bold, second-best are underlined.}
    \scalebox{0.91}{
    \begin{tabular}{l | cccccc | ccc}
    \toprule
    \multirow{2}{*}{\textbf{Approach}} &\multicolumn{3}{c}{\textbf{\cellcolor{blue!15}Semantic Score}} &\multicolumn{3}{c}{\cellcolor{blue!15}\textbf{Quality Score}} &\multicolumn{3}{c}{\cellcolor{green!15}\textbf{Total}} \\
    \cmidrule(lr){2-4} \cmidrule(lr){5-7} \cmidrule(lr){8-10}
    &\cellcolor{blue!15}\textbf{5s} &\cellcolor{blue!15}\textbf{10s} &\cellcolor{blue!15}\textbf{15s} &\cellcolor{blue!15}\textbf{5s} &\cellcolor{blue!15}\textbf{10s} &\cellcolor{blue!15}\textbf{15s} &\cellcolor{green!15}\textbf{5s} &\cellcolor{green!15}\textbf{10s} &\cellcolor{green!15}\textbf{15s} \\
    \cmidrule(lr){1-1}\cmidrule(lr){2-2}\cmidrule(lr){3-3}\cmidrule(lr){4-4}\cmidrule(lr){5-5}\cmidrule(lr){6-6}\cmidrule(lr){7-7}\cmidrule(lr){8-8}\cmidrule(lr){9-9}\cmidrule(lr){10-10}
    RollingForcing~\cite{liu2025rolling} &0.835 &0.831 &\ul{0.814} &0.800 &0.798 &0.802 &0.807 &0.804 &0.804 \\
    SelfForcing~\cite{huang2025self} &0.839 &\ul{0.837} &0.808 &0.834 &0.817 &0.832 &0.835 &0.821 &0.827 \\
    CausVid~\cite{yin2025slow} &0.813 &0.809 &0.794 &\ul{0.842} &\ul{0.831} &0.837 &\ul{0.836} &\ul{0.827} &0.828 \\
    RewardForcing~\cite{lu2026reward} &\ul{0.847} &0.836 &0.804 &0.808 &0.807 &0.835 &0.816 &0.813 &0.829 \\
    LongLive~\cite{yang2025longlive} &0.829 &0.825 &0.812 &0.816 &0.813 &\ul{0.838} &0.819 &0.815 &\ul{0.832} \\
    \midrule
    TANGO (ours) &\textbf{0.867} &\textbf{0.856} &\textbf{0.860} &\textbf{0.868} &\textbf{0.864} &\textbf{0.864} &\textbf{0.868} &\textbf{0.862} &\textbf{0.863} \\
    \bottomrule
    \end{tabular}
}
    \label{table:video_length_ablations}
\end{table}
\endgroup

To evaluate the effects of terminal points avoidance with respect to the length of the generated trajectories, we computed the 16 evaluation aspects of VBench~\cite{huang2024vbench} on videos of three different lengths. We present in \cref{table:video_length_ablations} the combined scores for the semantic aspects, visual quality as well as overall score, for videos of $5$\,s, $10$\,s and $15$\,s. This indicates that avoiding terminal points during test time consistently improves performance across different video lengths -- our approach achieves state-of-the-art results on all lengths. This highlights that terminal points do not only appear after the generation of very long sequences. Instead, they can be introduced to the generated trajectory at any point in time, preventing the generative model from continuing it without exiting the learned manifold of real videos, and thus, generating unrealistic content. Moving away from such points, through our noise guided optimization process, enables the model to faithfully continue the generated trajectory.

\section{Ethical Considerations}

Our work introduces a test-time approach to improve the consistency of diffusion-based autoregressive video generation models without requiring additional training data. While this directly benefits creative applications and world-modeling~\cite{ding2025understanding}, similar to any generative approach, it also exhibits dual-use risks. In particular, malicious actors can employ the generated media to fabricate disinformation or facilitate fraudulent activities~\cite{miao2024t2vsafetybench}. Yet, our commitment to open science practices enables the consideration of noise guidance during the development of AI-generated video detectors, to mitigate the risks of misuse. Moreover, any test-time adaptation approach effectively trades off test-time compute for improved performance, increasing energy consumption during inference. Nevertheless, recent studies in multimodal generative modeling have shown that training-time scaling usually provides diminishing returns~\cite{muennighoff2023scaling}, while bigger models exhibit decreased inference efficiency~\cite{maliakel2025investigating}. Under this bigger frame, our test-time adaptation approach not only provides a tool for surpassing limitations imposed by data availability, but also serves as an additional tool in defining a golden ratio between performance and efficiency in large-scale deployments.

\section{Extended Qualitative Evaluation}

\begin{figure*}
    \centering
    \subfloat[Text-to-video sample by TANGO. The main subject faithfully follows the given prompt, without generating  an unrealistic or static sequence.] {\includeinkscape[width=0.9999\columnwidth]{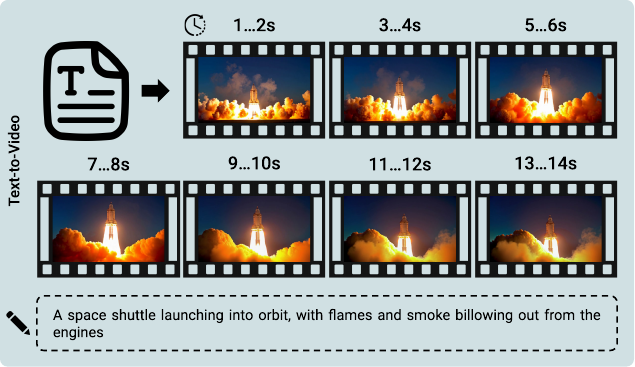}}\\ \vspace{1em} 
    \subfloat[Text-to-video sample by TANGO. The subject is naturally moving into the scene, performing the action described by the text prompt. Partial occlusions do not affect model's perception of the scene.] {\includeinkscape[width=0.9999\columnwidth]{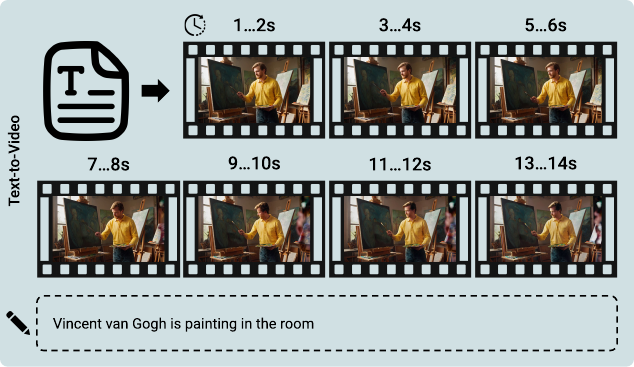}}
    \caption{Qualitative evaluation on text-to-video generation. (a) and (b) present 7 frames uniformly sampled from two generated videos across a time span of $14$\,s. On the bottom of each figure the employed text prompt is presented.}
    \label{fig:extended_qualitative_t2v_1}
\end{figure*}

\begin{figure*}
    \centering
    \subfloat[Text-to-video sample by TANGO. The video faithfully follows the provided prompt over time, maintaining an accurate representation of the background.] {\includeinkscape[width=0.9999\columnwidth]{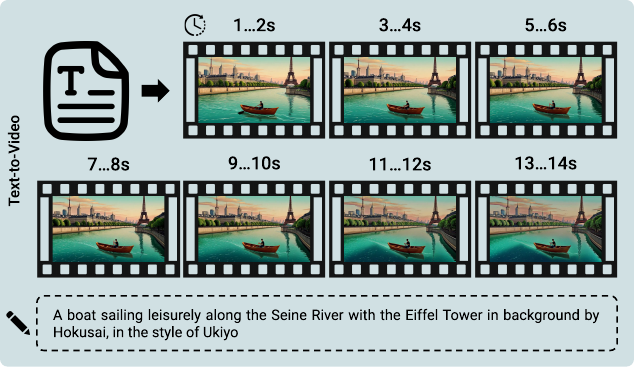}}\\ \vspace{1em} 
    \subfloat[Text-to-video sample by TANGO. The moving parts of the sewing machine are correctly identified, while the rest of the scene remains intact.] {\includeinkscape[width=0.9999\columnwidth]{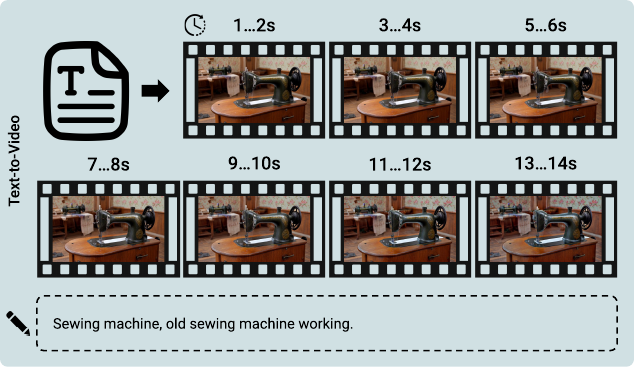}}
    
    \caption{Qualitative evaluation on text-to-video generation. (a) and (b) present 7 frames uniformly sampled from two generated videos across a time span of $14$\,s. On the bottom of each figure the employed text prompt is presented.}
    \label{fig:extended_qualitative_t2v_2}
\end{figure*}

\begin{figure*}
    \centering
    \subfloat[Image-to-video sample by TANGO. The environment is properly maintained, while the dancers move in a highly realistic manner.] {\includeinkscape[width=0.9999\columnwidth]{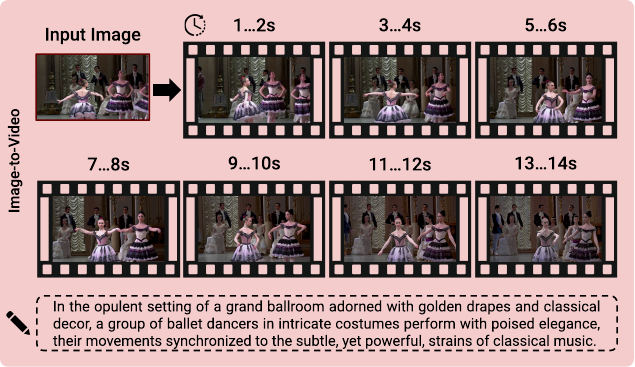}}\\ \vspace{1em} 
    \subfloat[Image-to-video sample by TANGO. The scene is naturally extended, while the main subject remains consistent and moves naturally.] {\includeinkscape[width=0.9999\columnwidth]{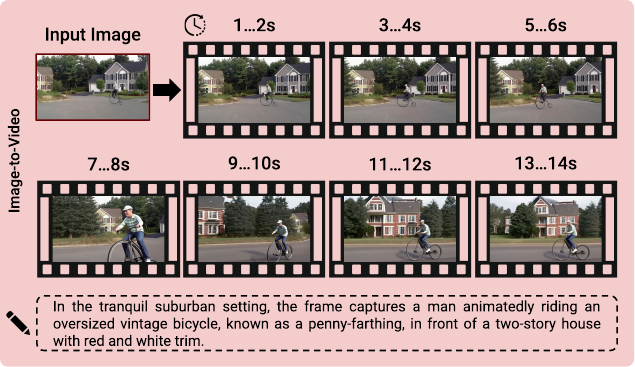}}
    \caption{Qualitative evaluation on image-to-video generation. (a) and (b) present 7 frames uniformly sampled from two generated videos across a time span of $14$\,s. On the bottom of each figure the employed text prompt is presented, while the input image is presented in the upper left corner.}
    \label{fig:extended_qualitative_i2v_1}
\end{figure*}

\begin{figure*}
    \centering
    \subfloat[Image-to-video sample by TANGO. Realistic aesthetics in the environment are maintained, even when the initial view gets out of camera.] {\includeinkscape[width=0.9999\columnwidth]{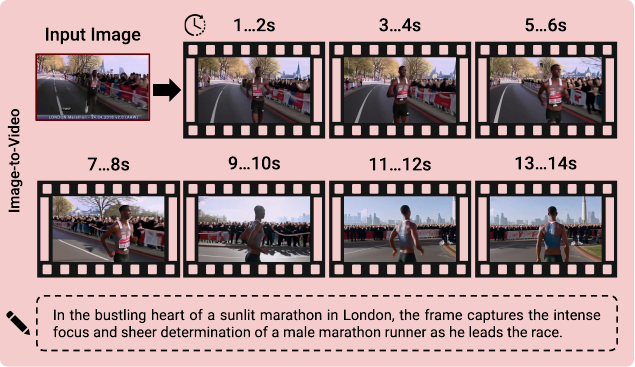}}\\ \vspace{1em} 
    \subfloat[Image-to-video sample by TANGO. People move naturally and realistic aesthetics in the environment are maintained.] {\includeinkscape[width=0.9999\columnwidth]{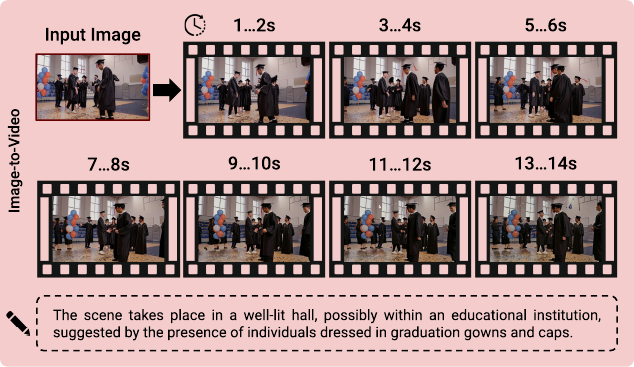}}
    \caption{Qualitative evaluation on image-to-video generation. (a) and (b) present 7 frames uniformly sampled from two generated videos across a time span of $14$\,s. On the bottom of each figure the employed text prompt is presented, while the input image is presented in the upper left corner.}
    \label{fig:extended_qualitative_i2v_2}
\end{figure*}

\begin{figure*}
    \centering
    \subfloat[Video-to-video sample by TANGO. The initial view is naturally extended and scene aesthetics are maintained.] {\includeinkscape[width=0.9999\columnwidth]{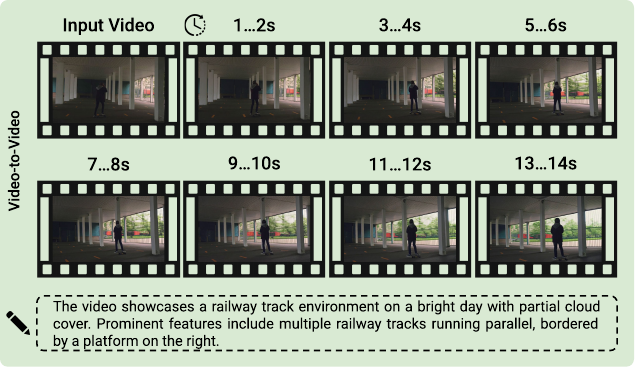}}\\ \vspace{1em} 
    \subfloat[Video-to-video sample by TANGO. The dancers move naturally, while their movements remain synchronized. ] {\includeinkscape[width=0.9999\columnwidth]{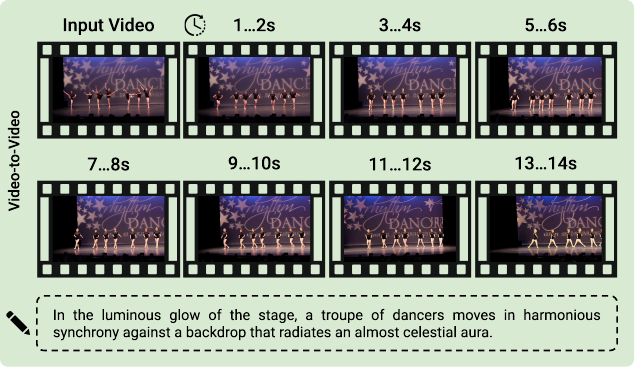}}
    \caption{Qualitative evaluation on video-to-video generation. (a) and (b) present 7 frames uniformly sampled from two generated videos across a time span of $14$\,s. On the bottom of each figure the employed text prompt is presented, while the input video is presented in the upper left corner.}
    \label{fig:extended_qualitative_v2v_1}
\end{figure*}

\begin{figure*}
    \centering
    \subfloat[Video-to-video sample by TANGO. The independent movement of the main subject and the camera is properly handled.] {\includeinkscape[width=0.9999\columnwidth]{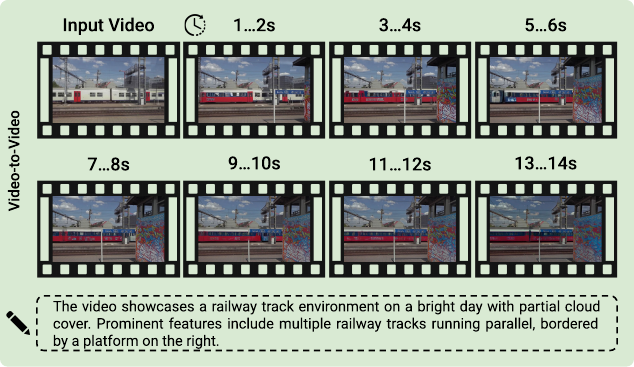}}\\ \vspace{1em} 
    \subfloat[Video-to-video sample by TANGO. Different views of the main objects are correctly generated, maintaining consistency along their movement and the camera's movement.] {\includeinkscape[width=0.9999\columnwidth]{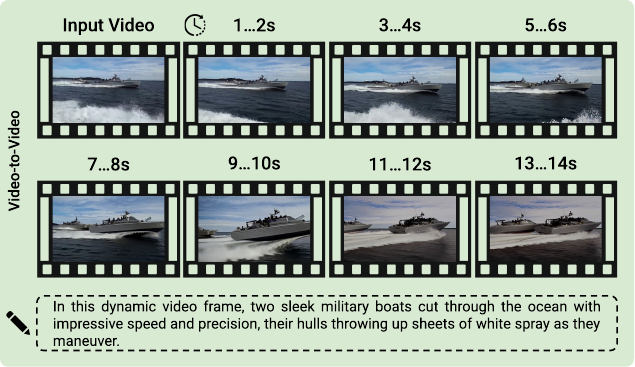}}
    \caption{Qualitative evaluation on video-to-video generation. (a) and (b) present 7 frames uniformly sampled from two generated videos across a time span of $14$\,s. On the bottom of each figure the employed text prompt is presented, while the input video is presented in the upper left corner.}
    \label{fig:extended_qualitative_v2v_2}
\end{figure*}

To visually assess the quality of the generated videos, we present several generated samples spanning text, image and video-to-video generation. In particular, under \cref{fig:extended_qualitative_t2v_1,fig:extended_qualitative_t2v_2,fig:extended_qualitative_i2v_1,fig:extended_qualitative_i2v_2,fig:extended_qualitative_v2v_1,fig:extended_qualitative_v2v_2}, we include seven frames from each sample, uniformly sampled across a $14$\,s time span. The employment of noise guided optimization to guide the generated trajectories away from terminal points improves the dynamic behavior of the depicted subjects, by enabling the generative model to generate video sequences it knows how to further continue.

\end{document}